%% file: conference_101719.tex
\documentclass[conference]{IEEEtran}
\IEEEoverridecommandlockouts
\usepackage{cite}
\usepackage{amsmath,amssymb,amsfonts}
\usepackage{algorithmic}
\usepackage{graphicx}
\usepackage{textcomp}
\usepackage{xcolor}

\usepackage{enumitem}
\usepackage{wrapfig}
\usepackage{hyperref}
\usepackage{multirow}
\usepackage{hyperref}
\usepackage{tabularx}
\usepackage{bm}
\usepackage{seqsplit}
\usepackage{stmaryrd}
\usepackage{booktabs}
\usepackage{fixltx2e}
\usepackage{xparse}
\usepackage[version=4]{mhchem}
\usepackage{footmisc}
\usepackage{tikz}
\usepackage{lipsum}

\DeclareGraphicsExtensions{.png,.pdf}
\label{packs}

\NewDocumentCommand{\heng}{ mO{} }{\textcolor{red}{\textsuperscript{\textit{Heng}}\textsf{\textbf{\small[#1]}}}}

\NewDocumentCommand{\chenkai}{ mO{} }{\textcolor{blue}{\textsuperscript{\textit{Chenkai}}\textsf{\textbf{\small[#1]}}}}

\NewDocumentCommand{\cheng}{ mO{} }{\textcolor{purple}{\textsuperscript{\textit{Cheng}}\textsf{\textbf{\small[#1]}}}}

\NewDocumentCommand{\jinfeng}{ mO{} }{\textcolor{cyan}{\textsuperscript{\textit{Jinfeng}}\textsf{\textbf{\small[#1]}}}}

\NewDocumentCommand{\weijiang}{ mO{} }{\textcolor{brown}{\textsuperscript{\textit{Weijiang}}\textsf{\textbf{\small[#1]}}}}

\NewDocumentCommand{\bluetext}{mO{}}{\textcolor{blue}{#1}}
\NewDocumentCommand{\cyantext}{mO{}}{\textcolor{cyan}{#1}}

\def\BibTeX{{\rm B\kern-.05em{\sc i\kern-.025em b}\kern-.08em
    T\kern-.1667em\lower.7ex\hbox{E}\kern-.125emX}}
\begin{document}
\onecolumn
© 2021 IEEE. Personal use of this material is permitted. Permission
from IEEE must be obtained for all other uses, in any current or future
media, including reprinting/republishing this material for advertising or
promotional purposes, creating new collective works, for resale or
redistribution to servers or lists, or reuse of any copyrighted
component of this work in other works.
\twocolumn

\title{Fine-Grained Chemical Entity Typing with Multimodal Knowledge Representation}

\author{\IEEEauthorblockN{Chenkai Sun, Weijiang Li, Jinfeng Xiao, Nikolaus Nova Parulian, ChengXiang Zhai, Heng Ji}\IEEEauthorblockA{\textit{Dept. of Computer Science}\\ \textit{University of Illinois at Urbana-Champaign} \\\textit{Champaign, IL, USA} \\
\{chenkai5, wl13, jxiao13, nnp2, czhai, hengji\}@illinois.edu}
}

\maketitle

\input{0abstract}


\input{1intro}

\input{3dataset}

\input{4method}
\input{5experiment}

\input{2related}

\input{6conclusion}
\bibliographystyle{IEEEtran}
\bibliography{IEEEabrv,custom}




\vspace{12pt}

\end{document}

%% file: 0abstract.tex
\begin{abstract}



Automated knowledge discovery from trending chemical literature is essential for more efficient biomedical research. How to extract detailed knowledge about chemical reactions from the core chemistry literature is a new emerging challenge that has not been well studied. In this paper, 
we study the new problem of fine-grained chemical entity typing, which 
poses interesting new challenges especially because of the complex name mentions frequently occurring in chemistry literature and graphic representation of entities. We introduce a new benchmark data set (CHEMET) to facilitate the study of the new task and
propose a novel multi-modal representation learning framework to solve the problem of fine-grained chemical entity typing by leveraging external resources with chemical structures and using cross-modal attention to learn effective representation of text in the chemistry domain. Experiment results show that the proposed framework outperforms multiple state-of-the-art methods.\footnote{The programs, data and resources will be made publicly available for research purpose.}





\end{abstract}

%% file: 1intro.tex


\section{Introduction}
\label{intro}

As the amount of research literature is growing exponentially, accurate and efficient information extraction (IE) methods are crucial for many downstream applications including question answering and knowledge reasoning.  Indeed, in scientific domains, IE models, including entity, relation, and event extraction, have already been widely developed for biomedical context~\cite{biomed_re, biomed_re4, biomed_ee,  biomed_ee2, biomed_jere}. 

However, one scientific domain largely overlooked by previous IE research is Chemistry. 
While chemistry research shapes the foundation of many biomedical studies, there has been so far little work done in extracting knowledge from core chemistry research literature. Previous work in Chemistry IE (ChemIE) mainly focuses on name tagging (e.g., recognizing chemical name spans)\cite{chemdner, chemdataextractor} or image~\cite{chemex}, and there is only one line of work we were able to find~\cite{chemu} on tasks other than name tagging (e.g., reaction event extraction). 

Unlike 
in most of the biomedical literature where 
entities are often expressed in natural language (e.g., water, aspirin), in chemistry, the entities are often complex formula-like names (e.g., 5,6-dihydroxycyclohexa-1,3-diene-1-carboxylic acid, \ce{H2O}). Research papers in chemistry contain many such mentions of chemicals and their reactions (for an example, see Figure~\ref{fig:framework}). These entity mentions are harder to be understood by existing language models because such complex names are created often to reflect chemical structure, instead of following morphological structure like other commonly used words.

Furthermore, many chemicals simply have never been coined with any nomenclature in natural language. The chemical mentions are essentially rare terms that cannot be learned well by only language models. Yet, extracting and typing such complex chemical entities is essential for virtually all the important downstream applications. 
If a comprehensive chemistry knowledge base can be efficiently constructed, chemicals can be discovered at a faster pace since models can learn from existing reactions to infer never-imagined ones, thus benefiting downstream applications such as those in biomedical research and chemical engineering industry.  

One fundamental building-block of IE 
is fine-grained entity typing (FET), which is the task of classifying each entity mention into a subset of pre-defined hierarchical classes 
(e.g., Organic Chemistry/Organic Compounds/Organonitrogen Compounds/Amides). Compared with prior IE work in Chemistry~\cite{chemdataextractor, chemdner, chemex, chemu}, FET enables more detailed representation of knowledge. The output of FET 
can also be used to improve performance on downstream tasks such as event extraction~\cite{biomed_ee} and relation extraction~\cite{biomed_re}.
Although 
much work has been done on FET 
in the news domain~\cite{label_bias, fetel, lin2019attentive}, no previous work has studied FET in the domain of chemistry literature.
In this paper, we conduct the first study of FET in chemistry literature. 
To facilitate the study of this new task, we construct CHEMET, the first dataset for fine-grained typing in the chemistry literature domain, for which we utilize external database for ontology construction, distant supervision for training data annotation, and human annotation for evaluation data labeling. 
The corpus consists of 100 open access papers from PubChem \footnote{\url{https://pubchem.1ncbi.nlm.nih.gov/}}, one of the mostly used chemistry knowledge base, on Suzuki-Coupling\footnote{\url{https://en.wikipedia.org/wiki/Suzuki_reaction}} theme (our techniques, however, are not limited to suzuki coupling text). One thing to be noted is that the task is different from classifying entries in PubChem chemical database as the mention from corpus might or might not exist PubChem. 

\begin{figure}
 	\vskip 0.2in
	\begin{center}
		\centerline{\includegraphics[width=0.7 
			\columnwidth]{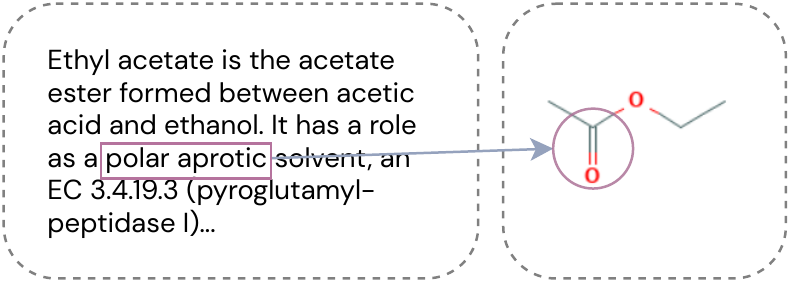}}
		\caption{The example shows multimedia definition (description text and chemical structure) from chemical database about ``Ethyl Acetate''. The circled sub-structure 
		often indicates the "polar aprotic" property of chemicals.}
		\label{fig:intuition}
	\end{center}
 	\vskip -0.2in
\end{figure}

FET is particularly challenging for chemistry articles, where domain-specific knowledge is heavily required to understand the text, making it an interesting problem to study from the perspective of Natural Language Processing (NLP) research. For instance, effective FET requires understanding of the reaction mechanism in the literature described by both equation image and text (about experiment conditions). Since a reaction is based upon chemical compounds, it additionally 
requires any FET method to capture the knowledge about the chemical entity mentions as well. 
However, many mentions are not explained or defined in local context, thus it would be difficult to decide the type of the mentions if we only use the sentence itself. To address this challenge, we propose to leverage the external knowledge resources in the chemistry domain, in particular, the PubChem database, where there are text descriptions and graphical representations of many chemical entities. 
We show an example entry of the PubChem database in Figure~\ref{fig:intuition}, where the description text and chemical structure are about the chemical ``Ethyl Acetate'' and some substructures are correlated with phrases in the description. 
Since the chemical structure is paired with a text description, which in general has strong semantic associations with the labels, we can leverage not only the external text information, but also the structural information, especially for entity type disambiguation. 

While existing work has often leveraged external knowledge resources in FET (e.g.,~\cite{fetel}), the exploitation of the chemical knowledge base (CKB) poses a new challenge on how to incorporate the multimodal information in CKB (i.e., both textual and graphical information). The knowledge bases in other domains that have been studied so far tend to be in the form of semantically well-defined graphs with text annotations.  To address this new challenge in effective use of an external KB with multimodal definition of entities, we propose a 
deep learning-based method that uses cross-modal attention to embed the structure and description text of chemicals into a common space as core features for classification.
As we will show in the experimental results, while existing language models may face challenges in understanding the chemical mention purely based on its surface form and contextual representation, the proposed multimodal representation learning method enables better understanding of the entity through its multimedia representation such as natural language description about its properties and its structure (or graph).

In sum, our work represents the first study of Fine-Grained Chemical Entity Typing in order to obtain a more detailed representation of knowledge buried in the chemistry literature, a largely under-explored yet promising field that has a great need for information extraction methods. The paper makes the following specific contributions:

\begin{itemize}
    \item We construct the first human-annotated dataset in fine-grained chemical entity typing and will release it to the public. Introducing FET in Chemistry enables efficiently classifying new molecules that are not possible with any existing data set.
    \item We propose and evaluate a novel method that utilizes multimodal knowledge to enrich entity mention representation. The multimodal component of our model is based on chemical structure and text alignment, which has never been explored before and can be applied to a variety of ChemIE tasks such as relation extraction and event extraction. 
    \item Using experiments on the dataset, we show that the proposed approach is effective at leveraging multimodal external knowledge resources and outperforms the state-of-the-art entity typing models. 
\end{itemize}




%% file: 3dataset.tex
\begin{table}
	\caption{Dataset Statistics for CHEMET}
	\centering
	\begin{tabular}{lllll}
		
		\toprule
		Setting &Anno.& \#Doc.& \#Sent. & \#Mention   \\
		\midrule
		Train &Distant&    80&6565&10318\\
		Dev &Human&20&520&1118\\
		Test&Human&            20&663&1295 \\
		\bottomrule
	\end{tabular}
	\label{datastats}
\end{table}


\begin{figure*}[ht]
	\begin{center}
		\centerline{\includegraphics[width=1.4
			\columnwidth]{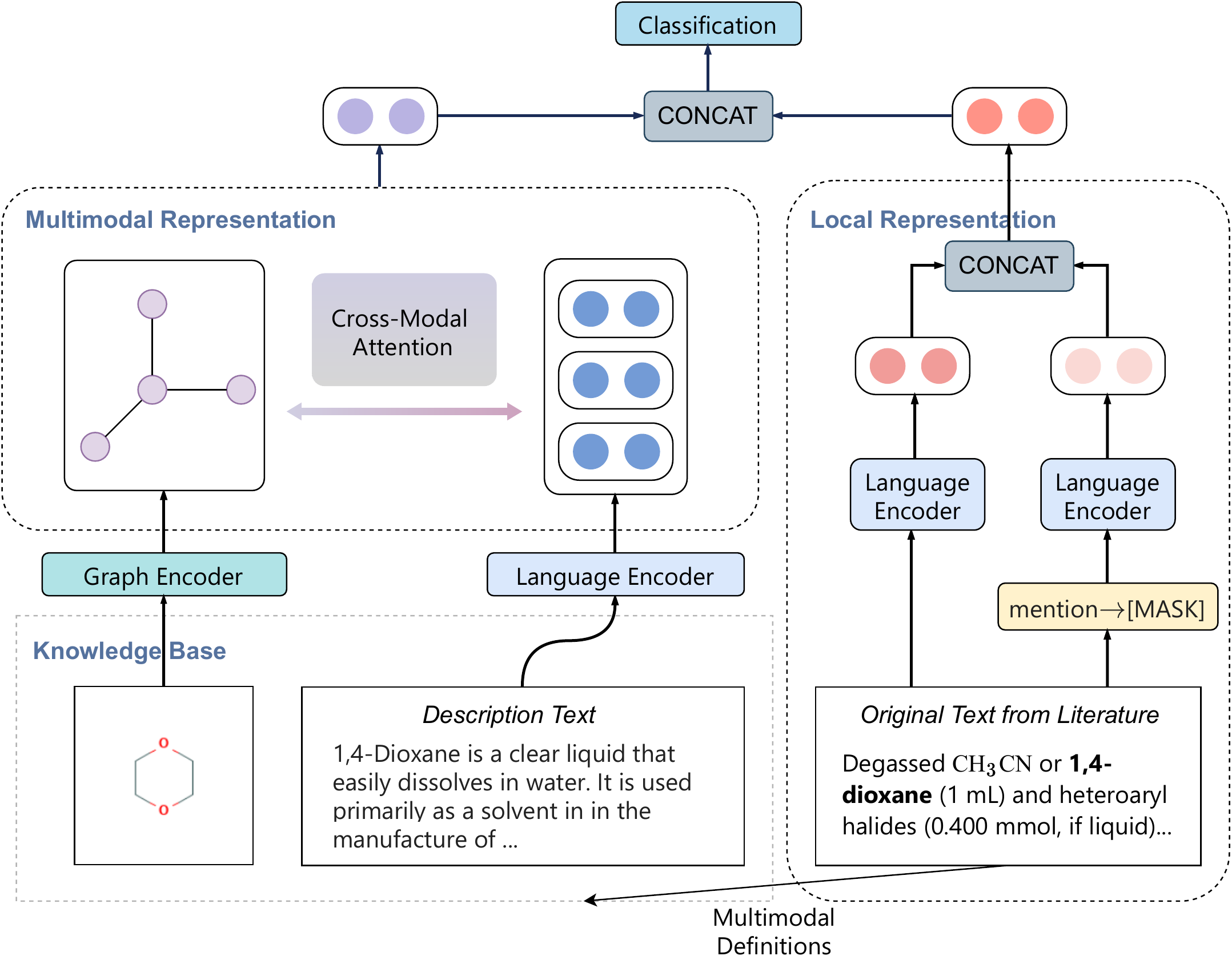}}
		\caption{Our fine-grained chemical entity typing model architecture. Please refer to Section~\ref{sec:method} for details. 
		}
		\label{fig:framework}
	\end{center}
	\vskip -0.2in
\end{figure*}
\section{Dataset}
\label{datacollection}




Since the problem we address has not been studied previously, our first challenge is how to evaluate the new task since there is no dataset available for fine-grained chemical entity typing. To this end, we have collected and annotated a new dataset, CHEMET, based on a corpus of 100 papers from PubChem with Suzuki-Coupling (a popular reaction mechanism) theme; the theme was chosen to align with chemistry annotators' domain knowledge. We will discuss the steps taken to construct the dataset below.

\noindent \textbf{Ontology Construction}.
In ontology construction, we focus on collecting the types that belong to chemicals commonly occurring in Suzuki-Coupling literature. We carefully select sub-categories from Wikipedia chemistry category page~\footnote{\url{https://en.wikipedia.org/wiki/Category:Chemistry}} as a fine-grained ontology. The reason for using Wiki ontology is that the only currently available comprehensive ontology for chemical compounds. Other ontologies are rather specific and cannot cover all entities of our interest. For example, ChEBI\footnote{\url{https://www.ebi.ac.uk/chebi/}} focuses on small molecules of biological interest, which is a subset of the compound types we are interested in. The decision was further validated after consulting domain experts. Moreover, Wiki ontology is not restricted to chemical compounds, and methods developed on Wiki ontology can be easily extended to other domains without requiring another domain-specific ontology that may or may not exist. One can easily obtain the Wiki ontology for another domain by changing the root of the DFS search as we described later in this section to the other target domain.


We define the following elements in our fine-grained ontology. \textit{Nodes} are Wikipedia \textit{categories} that form a \textit{tree} structure. Note that there are much more categories in the \textit{full Wikipedia ontology} than ours, and the full Wikipedia ontology is neither a tree nor a directed acyclic graph. There is a special kind of nodes in our ontology that are not Wikipedia categories. Those nodes have names starting with ``Other'', and are included in our ontology to handle entities that should belong to the parent node but not any siblings. \textit{Fine-grained types} are paths going from the root node to leaf nodes. Each node contains a bag of \textit{entities}, and the entities in a leaf node have the \textit{type} ending with that leaf node.

As an example, ``Chemistry $ \rightarrow$ Organic Chemistry $\rightarrow$ Organic Compounds $\rightarrow$ Hydrocarbons $\rightarrow$ Other Hydrocarbons'' is a fined-grained type with five nodes. Because the leaf node starts with ``Other'', it is not a Wikipedia category, but it contains entities that are Hydrocarbons but not Alkanes, Alkenes or Alkynes.





\noindent \textbf{Distant Label Generation}. We use Wikipedia and PubChem to generate distantly supervised labels for typed entities in text to enable noisy training and facilitate human annotation.
Each Wikipedia category can have sub-categories and associated pages. Starting from each node in our ontology, we trace its sub-categories in the full Wikipedia ontology recursively by performing a depth-first search (DFS) with a maximum depth of three. In the DFS we skip those sub-categories that are probably irrelevant to Chemistry. We use a spell-checker dictionary \cite{azman2012chemistry} with over 104,000 technical chemistry terms, and drop a category from the search if less than 20\% of the 1-grams in its name and the names of all its direct children are covered by the dictionary. The maximal depth of three and the threshold of 20\% are selected empirically by manually checking the quality of the results for some nodes. After the DFS, we use the Wikipedia page titles associated with any sub-category of a node as entities to populate that node. We further expand each entity with its synonyms from the PubChem database\footnote{https://ftp.ncbi.nlm.nih.gov/pubchem/Compound/Extras/CID-Synonym-filtered.gz}. For each node starting with ``Other'', we take the set difference between the entities in its parent node and those in its siblings. We now have produced a \textit{typed entity dictionary} that lists all fine-grained types and entities in the corresponding leaf nodes. Finally we use AutoNER \cite{DBLP:conf/emnlp/ShangLGRR018} with the typed entity dictionary to generate the distant labels for entities in the training and testing corpora. Here we give one example of such a distantly labeled sentence: To a solution of 2-bromo-5-(trifluoromethyl) $[\text{benzoic acid}]_\text{Carboxylic Acids}$ (5.00 g, 18.58 mol) in $[\text{methanol}]_\text{Other Functional Groups}$ (20 mL) $[\text{thionyl chloride}]_\text{Inorganic Compounds}$ (1.0 mL, 13.7 mmol) was added and the resulting mixture was refluxed for 8 h. 



\noindent \textbf{Human Annotation}. We hired five chemistry undergraduate students as our annotators. Following the annotation guideline, annotators are instructed to identify and type spans in the assigned sample documents using Brat~\cite{brat} interface. We randomly select 22 test documents from the corpus for annotation. Each document is assigned to two students. We assign junior student annotators to do the first pass, and assign senior student annotators to do the second pass to ensure the accuracy and consistency of our result. We use $F1$ score to calculate the alignment between first and second pass. After computing the $F1$ score between the first and second pass, we discard the two documents that have the lowest $F1$ score, which are 0.111 and 0.256, and take the result from the second round as the final data. The average $F1$ score of the rest of the 20 documents is 0.79, and we have 1,269 annotated sentences in total as the resulting dataset. The (cleaned) dataset statistics are shown in Table~\ref{datastats}. 

The size of the data was mainly limited due to our constraints on resources. Unlike in the news domain, annotating chemicals is a highly difficult task even for chemists since there are more than 100 million existing chemicals in PubChem, with many of them from corpora not on record, so annotators often need to analyze the experiment contexts to determine the chemical types.
However, as we will show in our experiments, 
this data set is already quite useful in that it enabled us to 
observe clear differences between methods in the paper and demonstrate the effectiveness of exploiting the multimodal information (section~\ref{sec:experiment}). 

One thing to be noted is that since the training set is distantly annotated, there is no mention of ambiguity involved in the text (i.e., each distinct mention is typed same in each sentence), but our development and test sets are manually annotated, so each mention in different sentences may have different labels.




%

%% file: 4method.tex
\section{Method}
\label{sec:method}
From NLP perspective, an interesting unique challenge in understanding chemistry literature 
is that the mentions are often expressed in complex, unnatural forms, as shown in the snippet in Figure~\ref{fig:framework}. Our main idea for addressing the challenge is 
to leverage external databases that contain multimodal definitions about chemical entities, including both their chemical structure and natural language description.  
To learn an effective representation from such multimodal definitions, we 
propose a general deep learning-based model with cross-modal attention that can learn effectively from the multimodal definitiosn and feed potentially many different downstream tasks with the learned multimodal representation of an entity mention as additional features.  

The overall model architecture is presented in Figure~\ref{fig:framework}. Given a sentence $S$ marked with mentions, we first extract external information (chemical structure and natural language description) by linking to PubChem; we use its search API to fetch entity information given a mention name. We also use a modified version of $S$ that masks the entire mention name, to combat the issue of complex chemical name (\ref{sec:context}). We then proceed to extract features from local context and external multimodal definition. The multimodal features are passed through cross-attention stage (\ref{sec:cmsa}) to learn a unified representation. Finally we use all features learned to classify the mention. Note that the multimodal learning part of the network is quite general and thus can be combined with any other downstream applications involving chemical entities. 

\subsection{Original Text Embedding}
\label{sec:context}

Given the original sentence $S$, we first insert a marker symbol ``*'' at the start and end of the mention $m$ during preprocessing, following~\cite{atlop}; similar approach is also used in \cite{marker2}. The model first encodes the original sentence with SciBERT~\cite{scibert}, a Transformer~\cite{transformer} based language model pre-trained on biomedical text. Let $T=[t_1, t_2, ..., t_z]$ be the tokens in $S$ after tokenization (we implicitly assume the presence of [CLS] and [SEP] tokens and omit them for brevity), where $z$ is the number of tokens. Then we pass the tokens into SciBERT to obtain contextual representations:
\begin{equation}
[\mathbf{t}_1, \mathbf{t}_2, ..., \mathbf{t}_z]=\text{SciBERT}([t_1, t_2, ..., t_z])\end{equation}

\noindent Where $\mathbf{T}=[\mathbf{t}_1, \mathbf{t}_2, ..., \mathbf{t}_z] \in \mathbb{R}^d$ and $d$ is the
number of hidden dimensions. We then use the embedding of ``*'' before $m$ as the mention embedding. Let us denote the mention embedding as $\mathbf{m}$.



\subsubsection{Context-only Embedding}
\label{sec:context_only}
Since chemical entities often involve complex names that are difficult to be understood, we also produce a representation that relies less on the word structure of the mention, since the mention does not often follow  morphological rules (e.g., [3H]MK-801, NSC-406186, 8-azido-[alpha-32P]ATP). We first replace the entire span of mention by [MASK], then the modified sentence is embedded by SciBERT and the embedding for the [MASK] token is used as the corresponding context-only embedding for the mention, denoted $\mathbf{m}_\text{MASK}$. The context-focused embedding is then concatenated with mention embedding to represent local information, denoted by $\mathbf{m}_\text{L}=[\mathbf{m};\mathbf{m}_\text{MASK}]$.

\subsection{Multimodal Encoder}
\label{sec:cmsa}


As one of our core contributions, we propose to incorporate multimodal definition to expand chemical representation, and to combat the difficulty of understanding complex chemical mention name (e.g., (E)-3-(3,4-dihydroxyphenyl)prop-2-enoic acid) purely based on context and morphological structure. 

Specifically, we use API provided by PubChem as the entity linker to retrieve chemical structure and natural language description for each chemical mention. Chemical structure refers to a graph where bonds are edges and atoms are nodes, and description text discusses chemical properties (e.g., Aspirin is an orally administered non-steroidal antiinflammatory agent). 

In order to learn concepts from multiple modalities that better correlate with target label and build more accurate representation of a molecule, we make use of the successful attention mechanism~\cite{transformer} to co-embed concepts (or molecule property) in text and substructure in chemical graph in order to capture interaction between different modalities. 
Formally, let $G=(V,E)$ denote the chemical graph with $a$ nodes, and $D=[ d_1,d_2,...,d_b]$ denote the sequence of $b$ tokens after tokenizing the description sentences. Similar to the embedding of the sentence from literature, we embed the tokens with SciBERT for which the output is $\mathbf{D}=[\mathbf{d}_1,\mathbf{d}_2,...,\mathbf{d}_b]$. We also embed the nodes in chemical structure using Graph Isomorphism Network (with edge features)~\cite{gin, gine}, a powerful graph neural network that can well capture different graph patterns. We randomly initialize embedding for each atom and bond type and use them to initialize node and edge embedding, and update node embeddings as follows:

\begin{equation}
\mathbf{n}_i^{l+1}=\text{FFNN}^{l+1}\bigg((1+\epsilon)\mathbf{n}_i^{l}+\sum_{j\in\mathcal{N}(i)}\mathbf{n}_j^{l}+\mathbf{e}^{l}_{j,i}\bigg)
\end{equation}

\noindent where $\mathbf{n}_i^{l}\in \mathbb{R}^d$ is the representation for node $i$ at $l$-th layer, $\epsilon$ is a tuning hyperparameter, $\mathcal{N}(i)$ is the set of neighbours of node $i$, and FFNN is a feed forward neural network with two hidden layers (the first one maps from $\mathbb{R}^d$ to $\mathbb{R}^{2d}$ with Tanh activation, and the second one maps $\mathbb{R}^{2d}$ back to $\mathbb{R}^{d}$ without activation). We denote node representation $\mathbf{N}=[\mathbf{n}_1,\mathbf{n}_2,...,\mathbf{n}_a]$.


We leverage self-attention mechanism of \cite{transformer} to learn dependency between different modalities. To achieve this, we first stack node and token embeddings as

\begin{equation}
\mathbf{X}=\begin{pmatrix}
\mathbf{N} \\
\mathbf{D}
\end{pmatrix}, \ \mathbf{X}\in\mathbb{R}^d
\end{equation}

\noindent where $\mathbf{X}\in\mathbb{R}^d$. Then the stacked embedding is passed through a Transformer layer to learn cross-modal association
\begin{equation}
[ \Tilde{\mathbf{n}}_1, \Tilde{\mathbf{n}}_2, ..., \Tilde{\mathbf{d}}_1, \Tilde{\mathbf{d}}_2...]=\text{Transformer}(\mathbf{X})\end{equation}




\noindent and the cross-modal feature is then obtained by mean pooling the output from the transformer layer

\begin{equation}
    \mathbf{f}_\text{cm}=\text{MeanPool}([ \Tilde{\mathbf{n}}_1, \Tilde{\mathbf{n}}_2, ..., \Tilde{\mathbf{d}}_1, \Tilde{\mathbf{d}}_2...])
\end{equation}

In addition, we preserve the unimodal graph representation by mean pooling over the node representation $\mathbf{N}$, to get $\mathbf{f}_{g}$. We also use the [CLS] token embedding $\mathbf{d}_{[CLS]}$ to represent unimodal text features. We then obtain a multimodal definition vector

\begin{equation}
    \mathbf{f}=[\mathbf{f}_\text{cm};\mathbf{f}_{g};\mathbf{d}_{[CLS]}]
\end{equation}
\subsection{Final Prediction}

Lastly, we predict the final entity type by using features from both local context and multimodal information
\begin{equation}
\mathbf{p}=\text{Sigmoid}(\text{FFNN}([\mathbf{m}_\text{L};\mathbf{f}]))\end{equation}

\noindent where $\text{FFNN}$ is a feed forward neural network mapping from $\mathbb{R}^d$ to $\mathbb{R}^{|\text{E}|}$, $E$ is the set of entity types, and $\mathbf{p}$ is the final probability distribution of classes.

\subsection{Training}
We use multi-label soft margin loss for training, that is,
\begin{equation}
\begin{split}
    \mathcal{L}=\frac{1}{C}\sum_{i=1}^{C}\Bigg( y_i\log\bigg(\frac{1}{1+e^{-x_i}}\bigg) +\\ (1-y_i)\log\bigg(\frac{e^{-x_i}}{1+e^{-x_i}}\bigg)\Bigg)
\end{split}
\end{equation}
\noindent In the equation, $C$ is the number of classes, $y_i$ indicates true (binary) label for class $i$ and $x_i$ is the predicted probability for class $i$.

%% file: 5experiment.tex
\begin{table*}[t]
	\caption{FET performance (\%) on CHEMET develpement and test set}
	\centering
			\begin{tabular}{cllll}
				\toprule
				\multirow{2}{*}{\textbf{Model}} 
				& \multicolumn{2}{c}{\textbf{Test}} 
				& \multicolumn{2}{c}{\textbf{Dev}} \\             
				&\textbf{Micro-F1}  & \textbf{Accuracy}& \textbf{Micro-F1} &\textbf{Accuracy}  \\  
				\midrule
				\cite{lin2019attentive} &  35.59&11.51 & 39.94 & 9.48 \\  \hline

				Bi-LSTM &  30.19 &  9.81 & 33.17 & 8.41 \\  
				SciBERT &  34.50 &  9.03 & 37.01 & 8.14 \\  
				SciBERT-C &  34.89 & 10.04 & 38.92 & 8.41 \\  
				SciBERT-CD & 37.19 & 11.89 & 39.29 & 11.54 \\      
				SciBERT-CG & 34.97 &  10.35 & 40.17 &  12.25 \\ 
				Our Model & \textbf{38.64}& \textbf{12.51}& \textbf{41.39} & \textbf{13.15} \\ 
				\bottomrule    
			\end{tabular}
		
	\label{table:result}
\end{table*}



		

\begin{table}[t]
\caption{Ablation Study (\%) on CHEMET develpement set}
	\begin{small}
	\centering
			\begin{tabular}{lll}
				\toprule
				\textbf{Model}
				&   \textbf{Micro-F1}& \textbf{Accuracy} \\  
				\midrule

				Full Model& 41.39 &13.15 \\\hline       
				w/o graph& 39.29 &11.54  \\   
				w/o description& 40.17 & 12.25 \\ 
				w/o cross-modal attention& 40.45 &11.45  \\       
				w/o context-only repr.& 40.76 &11.45  \\   
				\bottomrule    
			\end{tabular}

	\label{table:ablation}
	\end{small}
\end{table}
		
\section{Experiments}
\label{sec:experiment}
Since there are no other fine-grained chemical entity typing datasets, we evaluate fine-grained chemical entity typing on CHEMET. 
\subsection{Baseline Methods}
In the experiment, we compare our method with several baselines that are constructed based on SciBERT, which represent multiple ways of using the state-of-the-art methods for solving this new task. We also include a state-of-the-art baseline using label representation.

\noindent \textbf{Bi-LSTM}. A Bi-LSTM model that concatenates context embeddings (average of all token embeddings) and mention embeddings (average of mention token embeddings) with a linear layer in the end. To make the comparison fair, we use word vectors pre-trained on a 1 billion word chemical patent corpus as the pretrained word embedding\footnote{\url{https://github.com/zenanz/ChemPatentEmbeddings}\label{patent_url}}.

\noindent \textbf{SciBERT}. SciBERT~\cite{scibert} is a Transformer based language model pretrained on 1.14M papers from Semantic Scholar, in which 82\% are from the broad biomedical domain. A linear layer is applied on the embedding of the marker ``*'' embedding for classification. Based on SciBERT, we also implement \textbf{SciBERT-C}, which also uses context-only representation, referring to $\mathbf{m}_L$ in  section~\ref{sec:context_only}.

\noindent \textbf{SciBERT-CD} and \textbf{SciBERT-CG}. In addition to \textbf{SciBERT-C}, we create variants that additionally use either natural language description sentences (D) or chemical structure graph (G) for the chemical as features.


\noindent \textbf{Latent Type Representation}. \cite{lin2019attentive} uses a hybrid classification method beyond binary relevance to exploit type inter-dependency with latent type representation. Similar to Bi-LSTM, we use ELMo embeddings pretrained on 1 billion word chemical patent corpus\footref{patent_url} for a fair comparison.





\subsection{Implementation Detail}

Our model is implemented using PyTorch \cite{pytorch} and Huggingface Transformers \cite{huggingface} with SciBERT as text encoder. The model is trained on a single NVIDIA Tesla V100 GPU, taking approximately 5 minutes per epoch. The reproducibility and hyperparameter details can be found in google drive\footnote{\url{https://drive.google.com/drive/folders/1IMmDvsHb7a0JJfgnnhLmSX7gi_fOKsyJ?usp=sharing}}. 

\subsection{Result}
We use Micro-F1 and Accuracy (sample-average) as performance measures for the multi-label classification problem. Table~\ref{table:result} shows the overall results for the development and test sets on CHEMET. We can see that our model achieves higher performance than every other baseline on both metrics, even when 27\% of entities in the dataset are UNLINKABLE. We can also clearly see that the performance of SciBERT and SciBERT-C, which only use the local context, is not as good as other baselines which use the definitions of chemicals, showing the importance of incorporating non-local features, which may have added additional useful information
for entity type disambiguation. Finally the increase of the performance from SciBERT-CD and SciBERT-CG to our full model shows the benefit of learning a unified multimodal representation for chemical entities in chemistry literature. One may also find that the performance of baselines varies in different splits, which is mainly due to linkability and data distribution.

\subsection{Ablation Study}
To show the improvement made by each of the sub-modules in our method, we perform an ablation study on CHEMET development set and show the results in Table~\ref{table:ablation}. For the model [w/o graph] and [w/o description], we discard $\mathbf{f}_\text{cm}$ and correspondingly $\mathbf{f}_g$ or $\mathbf{d}_\textbf{[CLS]}$ in Section~\ref{sec:cmsa}. For the model [w/o cross-modal attention], we leave out $\mathbf{f}_\text{cm}$, and for the model [w/o context-only repr.], we keep every thing other than $\mathbf{\mathbf{m}_\text{MASK}}$~\ref{sec:context_only}. We see that taking off each sub-component of the model will lead to a decrease in performance, which demonstrates the importance of each part of the model.


	




\subsection{Error Analysis}

\begin{figure}[t]
	\centering
	\includegraphics[width=0.6\columnwidth]{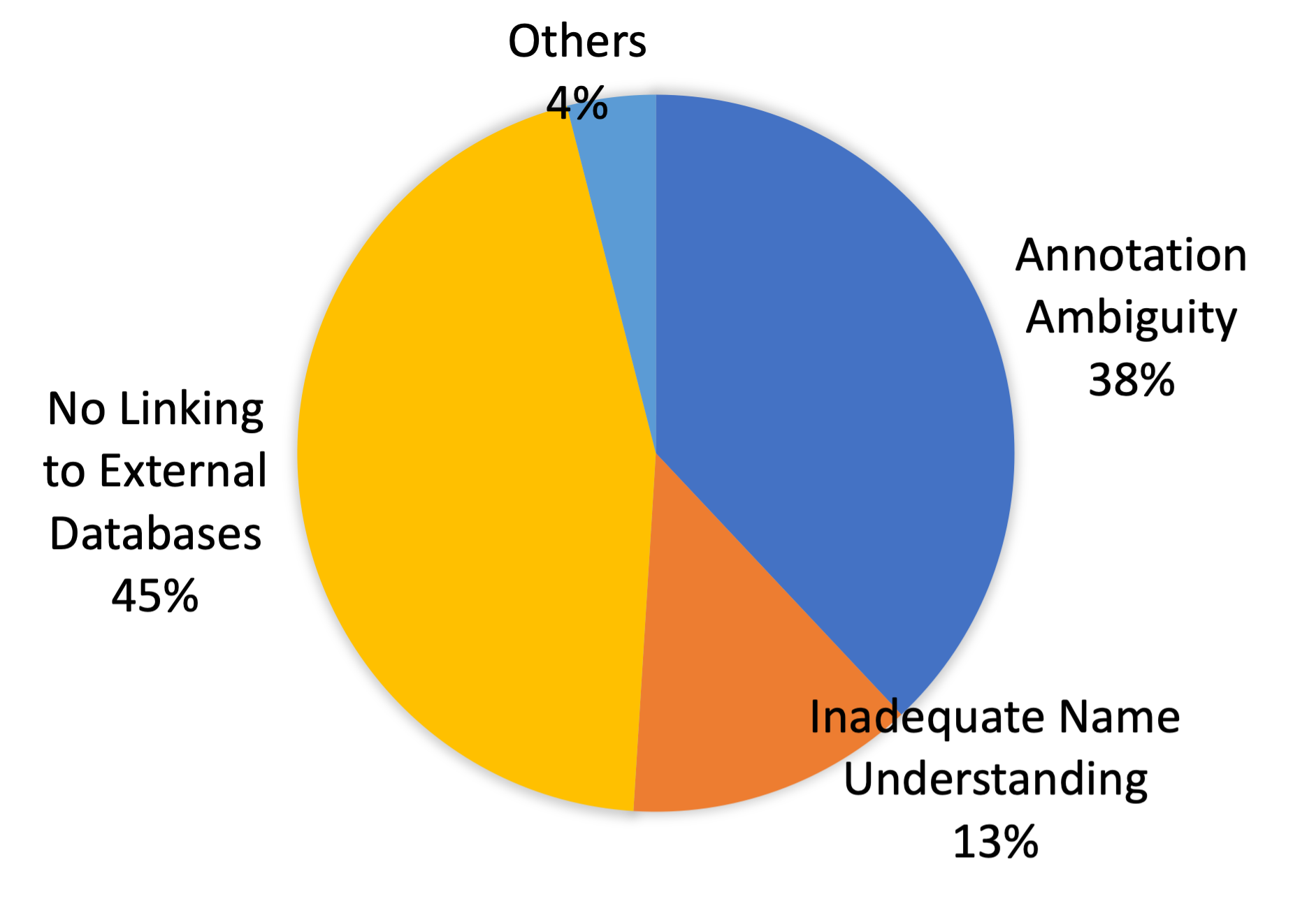}
	\caption{Distribution of remaining errors on the test set.}
	\label{error_dist}
	\vskip -0.2in
\end{figure}

We analyze 100 remaining errors and categorize them into four different cases  (shown in Figure~\ref{error_dist}). We discuss each of them below.

\noindent \textbf{No Linking to External Databases}. With the mechanism of our model, this is the most common cause of errors. Out of 100 cases we analyzed, there are 31 of them with such an issue. This may be due to the diversity of entities in chemistry literature, and a lot of these chemical compounds without a link to external databases are commercial compounds and synthesized chemical compounds. These are sometimes not all included in the PubChem database. In fact, over 27\% of mentions across dataset are not linkable, indicating that our model could have a significant improvement if we can fill the missing information, perhaps with an improved entity linker.


\noindent \textbf{Annotation Ambiguity}. This error occurs when the model correctly predicts all the true labels, but contains extra ``general'' label like Other\_Organic\_Compounds. 
For example, the mention EtOAc is predicted as Esters and Other\_Organic\_Compounds, compared to the ground truth Esters. This issue comes from the ambiguity during annotation. It might be nontrivial effort for annotators to see whether there is another existing tag for this chemical.  


\noindent \textbf{Inadequate Name Understanding}. 
This error occurs when the model only understands part of entity name. For instance, given the mention ``p-Me-Bn phenyl Pd(dppf)Cl2'', our model can only catch ``p-Me-Bn phenyl'' and correctly predict the entity to be an Other\_Aromatic\_Compounds (as one of the labels). However it doesn't recognize the ``Pd(dppf)Cl2'' part and thus misses other labels like Organometallic\_Compounds. The issue can be mitigated with more detailed context and description.




%% file: 2related.tex
\section{Related Work}




\subsection{Biomedical Information Extraction}
IE in biomedical domain has been widely explored~\cite{biomed_re, biomed_re4, biomed_ee,  biomed_ee2, biomed_jere}, but since it is a multi-disciplinary domain, it is crucial to develop effective IE models for subsidiaries. Our work targets assisting biomedical research by extending information extraction work in chemistry, aiming to automatically extract information in core chemistry research.
\subsection{Fine-Grained Entity Typing}
There has been a wave of Fine-Grained Entity Typing (FET) methods in recent years~\cite{label_bias, fetel, lin2019attentive}. \cite{label_bias} propose to capture label correlation by employing graph convolution network on label co-occurrence matrix. \cite{fetel} make 
use of an existing entity linker to obtain noisy external data in order to enrich and disambiguate mention representation. The authors also use entity linking scores as additional features. \cite{lin2019attentive} exploit type inter-dependency with latent type representation. Previous FET methods, however, only focus on news domain where text comes from news or wikipedia article and speech.  




\subsection{Multimodal Representation}

Multi-modal knowledge representation methods have been widely applied to tasks such as visual question answering and cross-modal retrieval between image and text. One line of deep-learning based alignment methods~\cite{clippaper, li2020gaia, manling_acl20} involves cross-modal alignment between separately learned word and image region representation. A recent popular line of research, including VisualBERT~\cite{visualbert} and VL-BERT\cite{vlbert}, integrates the reasoning process into pretraining, inspired from \cite{bert}. These models are fed with image-caption pairs and proceed to align regions and phrases by attention mechanism.

Different from the alignment among image, text, and audio, our method involves cross-modal attention between chemical structure and description text, which is a phenomenon specific to chemistry and has never been explored in previous work.

\subsection{Knowledge-Enhanced Language Representation}

Recently, there has been a lot of work~\cite{ erica,  kbert,kblstm} on incorporating external knowledge into language understanding. In~\cite{kbert}, triples are injected into the sentences as domain knowledge and attached to the tokens in the sentence. \cite{kblstm}, on the other hand, embeds words with KB concepts in an LSTM framework. 

As a unique contribution, our work is the first to draw a line between local context and external (chemical) entity structure information.

\subsection{Improve Information Extraction with External Knowledge}

There have been recent studies on improving information extraction with external data~\cite{fetel, Lai2021b, Zhang2021, Zhang2021b}, to help enrich or disambiguate local information. In \cite{Lai2021b}, the authors develop a model that aligns nodes in span graph and knowledge graph to learn a more distinctive concept embedding for joint biomedical entity and relation extraction. 
However, there has not been work that takes in physical structures of entities to enrich representation.

%% file: 6conclusion.tex
\section{Conclusions and Future Work}

In this work, we take the first step to explore a new task of fine-grained entity typing in chemistry domain and introduce a new dataset, CHEMET, to facilitate the study of the task. 
We also discuss the challenges in effective use of the multimodal definitions in this domain and 
propose a deep-learning based model that effectively incorporates multimodal definition of chemical mentions to improve the model's understanding on chemistry text. Evaluation results on CHEMET show
that the proposed model can effectively exploit multimodal definitions to improve performance for FET, outperforming all the baselines that we have compared with. Most importantly, while 27\% of entities are UNLINKABLE, we still achieve clearly better results. This not only proves the usefulness of our model but also shows its potential when an entity linking module can be incorporated (our next step).
While we have only evaluated it with FET, the proposed multimodal entity representation learning method is general and can be potentially applied to many other ChemIE tasks to enrich representation of chemical entities. A major remaining challenge is that many chemicals cannot be linked to the external database, either due to its varying mention form or because the database simply does not contain that particular entity (which is relatively more obvious for newer chemistry articles). In short, our contributions include not just a new task, a new data set, but also a new method for solving this problem that can serve as the baseline for future research. In the future, we will develop entity linking algorithm to not only match mention to databases better but also do cross-document linking (i.e., retrieve context for each entity from other documents).

\section{Ethics Consideration}
The human annotation is helped by five undergraduate students, who receive payment from the institution to conduct this research study. 
We assign 22 chemistry papers to student annotators based on their schedule, with the annotation mechanism that junior student annotator doing the first pass and senior student annotators doing the second pass. We instruct them to follow the annotation guideline, which includes an example of a completed annotation document, instructions to use the annotation tool, and four actions they need to take based on the pre-annotated documents. We ensure that 
every annotator understands the annotation task by assigning a test document and reviewing the result, then we supervise them virtually throughout the annotation process. We will publish this dataset to facilitate future research in this area.